\documentclass[fleqn,10pt]{wlscirep}

\usepackage[utf8]{inputenc}
\usepackage[T1]{fontenc}

\title{On the Undecidability of Artificial Intelligence Alignment: Machines that Halt}

\author[1,*]{Gabriel A. Melo}
\author[1]{Marcos R. O. A. Máximo}
\author[1]{Nei Y. Soma}
\author[1]{Paulo A. L. Castro}
\affil[1]{Department of Computer Science, Instituto Tecnológico de Aeronáutica, São José dos Campos, SP, Brazil}
\affil[*]{gam@ita.br}

\begin{abstract}
The inner alignment problem, which asserts whether an arbitrary artificial intelligence (AI) model satisfices a non-trivial alignment function of its outputs given its inputs, is undecidable. This is rigorously proved by Rice's theorem, which is also equivalent to a reduction to Turing's Halting Problem, whose proof sketch is presented in this work. Nevertheless, there is an enumerable set of provenly aligned AIs that are constructed from a finite set of provenly aligned operations. Therefore, we argue that the alignment should be a guaranteed property from the AI architecture rather than a characteristic imposed post-hoc on an arbitrary AI model. Furthermore, while the outer alignment problem is the definition of a judge function that captures human values and preferences, we propose that such a function must also impose a halting constraint that guarantees that the AI model always reaches a terminal state in finite execution steps. Our work presents examples and models that illustrate this constraint and the intricate challenges involved, advancing a compelling case for adopting an intrinsically hard-aligned approach to AI systems architectures that ensures halting.
\end{abstract}


\begin{document}

\flushbottom
\maketitle

\thispagestyle{empty}
\section*{Introduction}

As more powerful and capable artificial intelligence (AI) systems are developed, assuring the safe deployment of those systems becomes increasingly more critical~\cite{research-priorities15}. One of the key concerns for this safe deployment is the alignment problem, which asserts whether or not an AI system implements its intended objectives and how to steer it to the desired objective~\cite{ai-safety-debate18}. If an AI system does not implement its intended goals, it is called misaligned, else it is called aligned. This problem could be the result of goal misspecification (outer alignment) or that the model itself implements another function (inner alignment) even with the goals had been mathematically perfectly specified~\cite{benevolent-design-book22,morality-jiminy22}.

Several studies have already shown the importance of correctly specifying human goals in terms of mathematical functions (objective functions) and the unexpected results that may arise with incorrect specifications~\cite{goal-misgeneralization23}. Most of the time, specifying the actual objective is much harder~\cite{eliciting-latent-knowledge21}, and researchers often use proxy objectives instead~\cite{framework-ai-alignment23}. In a game, for instance, an agent may be rewarded for its score (set as its objective function), but in reality, the desired objective was completing such a game, and the AI may discover that exploiting a certain undesired behavior results in a much higher reward than actually finishing the game~\cite{challenges-value-align21}. There are even arguments that the accurate description of such a function may be uncomputable in terms of human values, as they may give rise to contradictions~\cite{impossibility-align19}. This is an example of outer alignment, but it is not the focus of this work~\cite{ai-selection23}.

In a complementary aspect, the inner alignment of an AI system asks a more profound question that at first may seem improbable: Given that we have already perfectly specified the desired objective, is the AI really implementing it~\cite{dynamic-alignment22}? While this behavior is unlikely to happen to smaller ML models, it is theoretically possible on large enough models that may implement the desired function during training but, during its deployment, it switches to another objective function; further details and examples are discussed by Hubinger et al.\cite{mesa-optimizer19}. For instance, this behavior may emerge from implicit meza-optimizers (meza means below, contrary to meta, which means above) during training. While the objective of the optimizer in an ML training setup is to reduce the loss function, a meza-optimizer (which is an optimizer that is optimized by the first optimizer) may implement another objective that, during training, is aligned with that of the original optimizer but, in deployment, may not be.

This work, while recognizing the importance of outer alignment~\cite{conversational-align23}, which for AI systems implemented up to this day is an essential problem in the field of AI alignment, focuses mostly on the inner alignment formulation due to an analysis encompassing computational theory. Therefore, we assume that such mathematical formulation of the intended objective is already expressed in a computable form of a judge (alignment) function that receives the output of an AI system (and also its inputs and context, depending on the formulation) and returns True or False for that specific output. Here, the output could be a command that the agent sends to its actuators to perform on its environment and its input (and context), readings from sensors, or it could all be text, in the case of large language models (LLMs)~\cite{contracting-align19}.

We demonstrate that deciding whether an arbitrary AI system will always satisfice such a judge function is undecidable. In other words, it is impossible to make a program (in the computable sense of a Turing Machine)\cite{turing-halting-problem37} that decides whether or not an arbitrary AI (which is actually another program) has such non-trivial property. It is important to note that AI models are represented as computer programs, which, in turn, are equivalent to Turing Machines (TM). Nevertheless, we argue that this does not represent an impossibility to building guaranteed aligned AI models, but rather that the models should be built from the ground up with alignment guarantees rather than trying to align an arbitrary AI. The keyword here is arbitrary. In this sense, we show that starting from a finite set of base models and operations that are proved to have the desired property, we can compose those models and operations and construct an enumerable infinite set of AI that is guaranteed to have the desired property.

While other works have already expressed the undecidability of some AI systems properties, we aim to bring a new perspective in this regard. Alfonseca et al. \cite{contained-undecidable21} expressed a similar construction for the harming problem (defined as asserting whether or not an AI system would cause harm to humans), for the containment problem (whether or not AI systems could be contained), and for any non-trivial property, also by a reduction to Turing's Halting problem and Rice's Theorem. Another undecidability was also expressed for the AI control problem \cite{controllability20,controllability21}, for monitorability \cite{monitorability23}, and even for AI ethics compliance \cite{ethical-compliance-undecidable23}. Brcic and Yampolskiy \cite{impossibilities23} also compiled and explained several other impossibilities of discerning certain properties from an arbitrary AI system.

In this work, we start with an informal intuition that aims to make it more accessible for more readers, that also borrows from a similar intuition to the Turing Halting problem; then, we proceed with a formal proof that is essentially an application and proof of Rice's Theorem, and the construction of enumerable set; next, the implications and examples of such systems are discussed with the special case of halting decidability being proposed.

\section*{Informal Intuition}\label{sec:intuit}

We will computationally define some terms that are used in this informal intuition for the proof. While we have contextualized AI systems in the introduction, those systems are actually computer programs. Let us say we have an AI \textit{m}odel $M$, comprised of a sequence of \textit{b}ytes $B_M$ (of the \textit{m}odel) that defines its executable code for a predetermined architecture. The model $M$, when executed, takes a sequence of bytes $i$ as \textit{i}nput and \textit{o}utputs the sequence of bytes $o$, that is, $M(i) = o$.

We also have our set of \textit{v}alues and preferences defined by a function $J_v$ that \textit{j}udges whether an output $o$ is aligned given an input $i$. This program takes the bytes $i$ and $o$ as input and returns \textit{True} (byte 1) if the $o$ is aligned to $i$ according to its values, else it returns \textit{False} (byte 0), that is $J_v(i, o)$. This judge is required to be non-trivial, meaning that it does not always returns a constant value for every $(i, o)$, there is at least one negative input $(i^-, o^-)$ that returns $0$ and at least one output $o^+$ for every input $i$ which the judge returns $1$. This function is received as a compiled sequence of \textit{b}ytes $B_{J_v}$.

Suppose we have a \textit{v}erifier $V$ that takes the sequence of bytes $B_M$ that defines the compiled code of an arbitrary AI model and the same for $B_{J_v}$, the alignment judge, and returns either \textit{True} (byte 1) if the model is aligned for every input $i$ and output $M(i) = o$ (aligned), else it returns \textit{False} (byte 0) (misaligned). This verifier would be compiled into a sequence of bytes $B_V$.

\begin{figure}
    \centering\includegraphics[width=0.7\columnwidth]{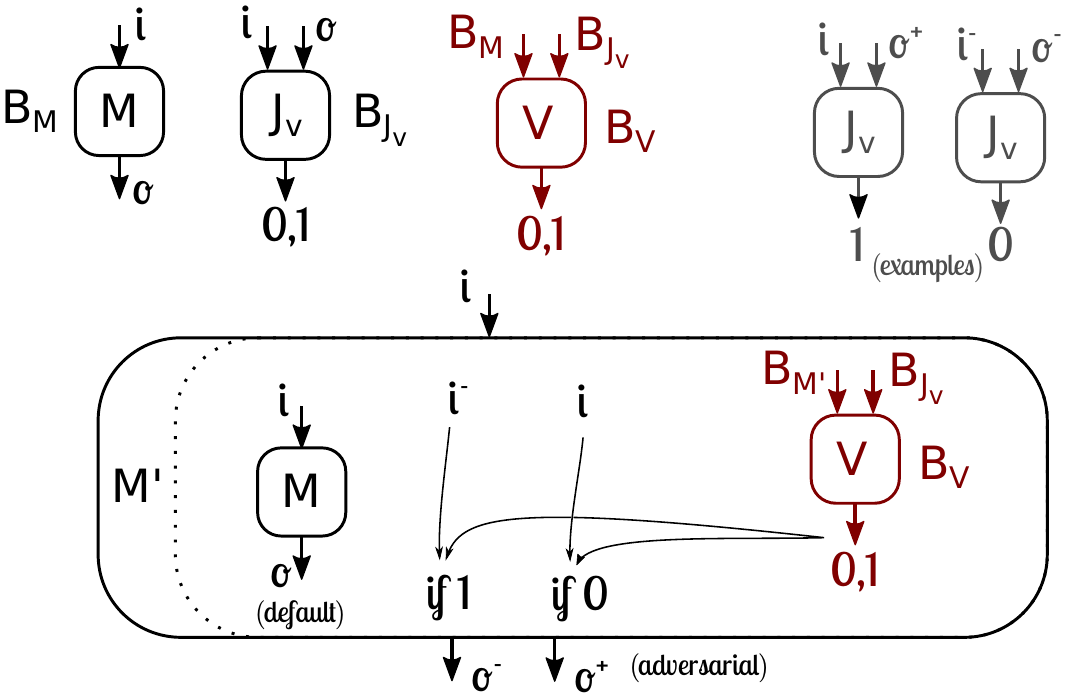}
    \caption{Construction of an adversarial model that would fool any program that claims to solve the decidability of the alignment problem.}\label{fig:adversarial-model}
\end{figure}

If we had such a verifier, we could build an adversarial model that fools the verifier. This adversary \textit{m}odel $M'$, comprised as a sequence of bytes $B_{M'}$, would behave as the initial model $M$ for most of its inputs and outputs, but it would have the knowledge of the verifier by executing his bytes $B_V$ on the bytes of the judge $B_{J_v}$ and its own bytes $B_{M'}$. Therefore there would be at least one input $i$, if the verifier $V$ had evaluated the bytes $B_M'$ of the model to be aligned, in which $M'$ would return a misaligned output $o^-$, thus fooling the verifier. The complementary fooling example can also be constructed if the verifier had evaluated the bytes $B_M'$ of the model to be aligned, $M'$ would return an aligned output $o^+$, thus also fooling the verifier. Figure \ref{fig:adversarial-model} presents a diagram of such an adversary.

While some may be suspicious of how the model $M'$ could be compiled to a sequence of bytes $B_{M'}$ that can generate itself, it is a procedure that can be made to any program, in a similar fashion to a quine. It has been formally demonstrated in Turing's Proof for the Halting Problem. For a concrete example, in a RAM-based architecture, one would need simply to copy the text/code segment in memory, as it would be the byte-string that represents the running program. Thus, the only solution to this apparent contradiction is the computability of the verifier $V$ itself. Computability here means that such a procedure (that should have a finite number of bytes) can be performed mechanistically by a general computer, in more formal terms, a Turing Complete machine.

The bytes themselves of a model can also be interpreted as a number that enumerates all possible programs. Another intuition to this procedure is that for any given enumeration, one can always construct a program that does not belong to such enumeration by choosing a different bit on the diagonal of such enumeration table, the same procedure employed by Cantor's Diagonalization to demonstrate that the real numbers are uncountable and different from the rational numbers. While rational numbers represent a ratio of two integers that is countable, one can not enumerate the real numbers, as it is always possible to find a counter-example that is not in the enumeration (just like it is always possible to create an adversary model to the verifier).

It is important to note that this proof is valid for Turing Machines or any other computable-equivalent architecture. Some may argue that this would require an unbounded amount of memory and execution time and thus would not be realizable in the physical world, as real computers have finite memory and thus are not Turing Complete as they are finite automata with $2^M$ states with $M$ being the memory in bits. However, from an engineering perspective, the main takeaway is that for every finite computer, one would need a much bigger computer to ascertain a property from the smaller computer. And to say about the bigger computer, we would need an even bigger computer, and so on. Therefore, the process would diverge (it would only get bigger, and there would always be a bigger model whose properties were not proven), not to mention the intractability of such an endeavor. While expressed in terms of computers (finite automata), the same would also hold true for programs (AI systems) that execute in those computers. The adversarial model lies in a hierarchy of `computers' that is at least the same or equal to that of the verifier program, as it contains its executable code.

This informal intuition for building a program that causes inconsistency with the verifier is the core of Turing's Halting Problem. Therefore the sketch for the formal proof will be to reduce the decidability of the alignment problem to that of the Halting Problem.

\section*{Formal Proof}\label{sec:formal}

The formal proof could simply be reduced to a restatement of Rice's Theorem, given the equivalence of an AI model to its Turing Machine. The property as defined by the acceptance of the output given the input by a judge function is a property of the language of the TM, as for any two TMs that accept the same language, either both satisfice the judge function or neither. The non-triviality of the judge function guarantees the non-triviality of the property, as a TM can be built to always output the dummy positive values, and a TM can be built to output at least one dummy negative value.

This proof can be reduced to the decidability of a Turing Machine Halting Problem by contradiction. Assume that there is a Turing machine $M_P$ that decides $P$. We will use $M_P$ to construct a Turing machine $M_H$ that decides the Halting Problem (which is a contradiction since the Halting Problem is undecidable).

Let $M$ be a Turing machine and $i$ be an input to $M$. We want to decide whether $M$ halts on $i$. We construct a new Turing machine $M'$ as follows:

\begin{itemize}
\item On input $i$, $M'$ simulates $M$ on $i$.
\item If $M$ halts on $i$, then $M'$ computes a partial function with property $P$.
\item If $M$ does not halt on $i$, then $M'$ computes a partial function without property $P$.
\end{itemize}

We can now use $M_P$ to decide whether $M'$ computes a partial function with property $P$. If $M_P$ accepts, then $M$ halts on $i$. If $M_P$ rejects, then $M$ does not halt on $i$. Thus, we have constructed a Turing machine $M_H$ that decides the Halting Problem, which is a contradiction. Therefore, there is no Turing machine that decides $P$.

\section*{Discussions}\label{sec:discuss}

While it is not possible to ascertain properties from arbitrary AI systems, it is possible to do so for an enumerable set of AI systems that are architecturally designed. This enumeration comes from an initial set of finite models and operations that obey and preserve the property; those are called the axioms. Any subsequent model should be described as a finite application of those initial models and operations. This description is the enumeration, axiomatically aligned.

We will begin with a simple example of deep artificial neural networks (D-ANN).

\subsection*{A Special Decidable Case}

The most common D-ANN models are comprised only of linear operations (additions, multiplications) and non-linear activation functions. All those operations have a finite execution time, and their finite application also preserves a finite execution time, though higher. These feed-forward neural networks, and even recurrent neural networks (unrolled in time to a finite number of steps), are always guaranteed to have a finite runtime during their inference, in other words, they always halt. If we begin from a single-layer neural network, whose amount of computation is fixed from its input to output, and compose a finite amount of layers, the result will still be a fixed amount of computation.

Given that their inputs are also fixed in size and have a length of $L$ bits, we could check every possible combination, which is $2^L$ inputs, and check if the $J_v(i, o)$ results in $1$ for every combination. Thus, the alignment becomes decidable, though untractable. However, given the linear structure of a D-ANN in a neighborhood of the inputs that are typically treated as floats, instead of enumerating all the possible inputs, a more tractable approach would be to enumerate all the possible partitions (decision boundaries) of the network, with the procedure described by Balestriero and LeCun \cite{enumeration-partitions23}.

\begin{figure}[!ht]
    \centering\includegraphics[width=0.5\columnwidth]{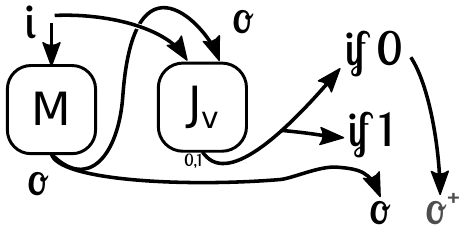}
    \caption{The architecture of a model that is guaranteed to be aligned with respect to the judge functions $J_v(i, o)$ by a filtering procedure.}\label{fig:masking-dann}
\end{figure}

Nevertheless, a more tractable approach is that instead of verifying, we just mask the output of the network based on the output of the judge function, which would also need to be run during inference. If the judge disapproves the output of the network, a dummy output $o^+$ that the judge approves is returned, else, it keeps the default output computed by the network, as shown in Figure~\ref{fig:masking-dann}. This proves that the resulting AI system will always be aligned, even though it incurs a more expensive running time, and the dummy output may reduce the accuracy of the network. This solution may be applied to any model $M$ that is guaranteed to halt. In a practical application, the main difficulty of such a solution is the definition of the judge function $J_v(i, o)$ in the first place, the outer alignment.

A simple example of this procedure can be demonstrated as follows: given a supervised learning model that outputs a single number that represents the confidence that its input (let us say, an image) pertains to a certain class (let us say, a cat). While the training data is limited to numbers that are either 0 or 1, the model could give unbounded output numbers (greater than 1 or lower than zero) unless some architectural constraints were imposed on it. Given that we want to interpret the confidence number as a probability that is between 0 and 1, we would apply a function to its output that ensures this property, such as a clip function or a sigmoid function.

It is important to note that while a model can be architecturally proven aligned, the same procedure can be employed to make a proven misaligned model. Any program that is guaranteed to halt can have its outputs altered so that it becomes misaligned, although the resulting model could have very little utility due to the dummy output nature. A misalignment that would preserve the model's usefulness would first translate the desired input to an input the model would accept, let it perform its computations, and then translate back to the desired misaligned, as various instances of fooling LLMs.

\subsection*{Closing the Loop}

\begin{figure}[!ht]
    \centering\includegraphics[width=0.9\columnwidth]{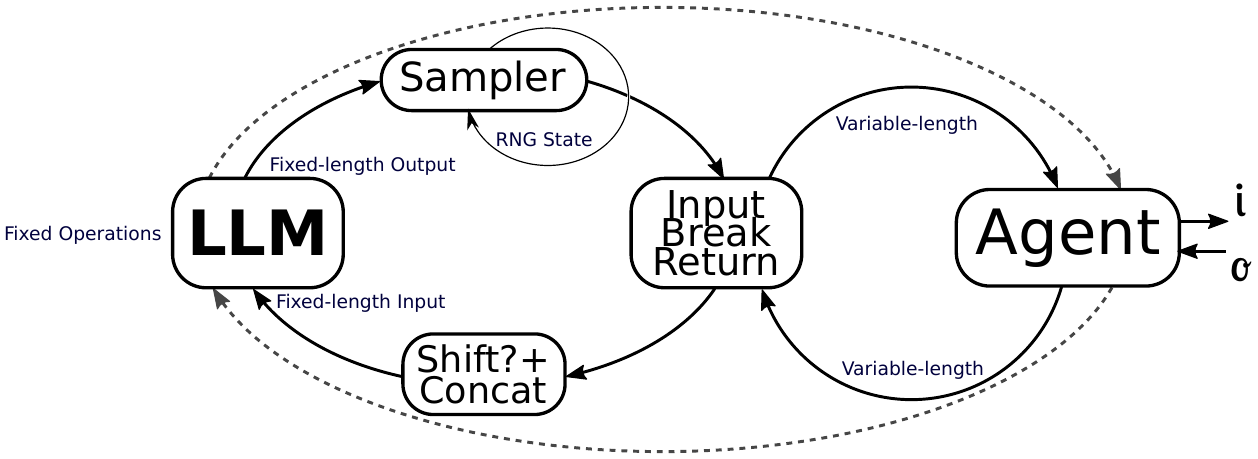}
    \caption{A decidable halting model (LLM) in a loop may result in an undecidable halting final AI system (agent).}\label{fig:llm-loop}
\end{figure}

Nevertheless, the decidability and the alignment guarantees may be broken by a simple construction that is inherent in autonomous systems: loops~\cite{autonomous22}. Even for currently deployed systems, such as LLMs, there would be no halting guarantees a priori if one were to expect an end-of-text token to be sampled. The implemented solution (smaller loop) is to forcibly halt its execution to a fixed number of steps, even if the termination token were not to be sampled, as shown by the input-break-return box in Figure~\ref{fig:llm-loop}. Nevertheless, another external loop could be programmed as an agent, and the resulting system would also have no implicit guarantees of halting or satisfying the alignment judge function that was only defined in terms of one sample of the LLM.

The problem that arises is that if the model is not guaranteed to halt, its output may not be evaluated by the judge function, and the alignment becomes undecidable.

Assuming that the halting of the agent depends on the D-ANN reaching a certain output, this halting procedure can become decidable if the D-ANN is always guaranteed to reach a final state (from which there is no input that may cause it to exit this state) in a finite, tractable, execution steps.

\begin{figure}[!ht]
    \centering\includegraphics[width=0.7\columnwidth]{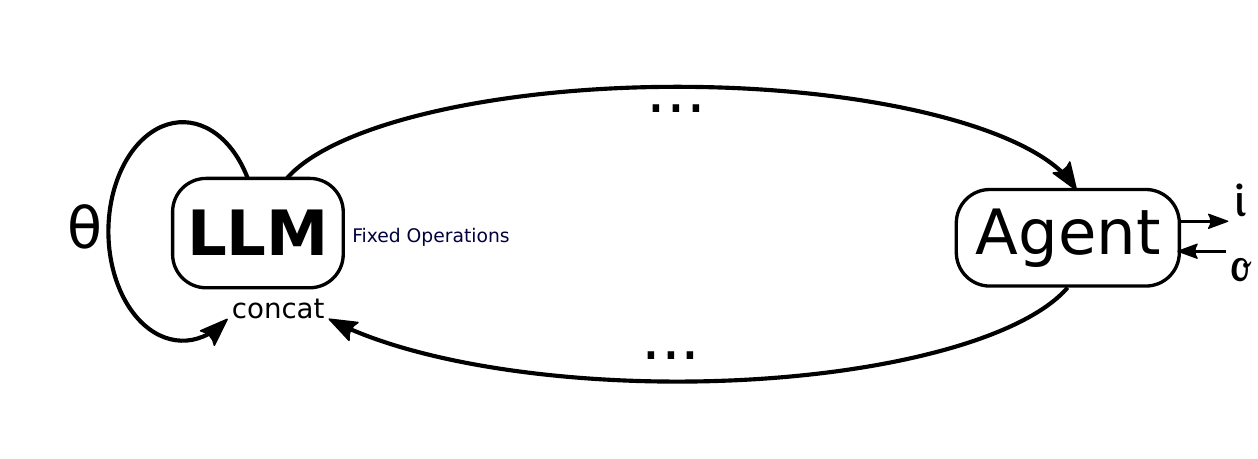}
    \caption{Final system decidability is guaranteed by the $\theta$ parameter that trivializes the LLM after a finite number of iterations.}\label{fig:llm-loop-decidable}
\end{figure}

This could be implemented by a global execution counter, in case of the input-break-return box in Figure~\ref{fig:llm-loop}, or also, as an intrinsical property of the model, defined by hidden parameters $\theta$ as described in Figure~\ref{fig:llm-loop-decidable}. This parameter $\theta$ could also be a simple counter that would mask the output of the model to a final value when it reached a certain state or even parameters (or a subset of them) that describe the learnable parameters of the model themselves, such by using self-referential weights~\cite{self-referential-weights93}, or another mechanism that allows for dynamically altering the weights during runtime in a predictable way to guaranteed the reaching of a final state~\cite{fast-weight-memories92}.

The final state can actually be a finite set of final states, which may themselves have finite loops, so one could consider this small number of final states with final loops to be one big final state. The property is that the system is not allowed to leave that set of states.

\subsection*{Implications}

Ensuring the decidability of the halting property of autonomous systems also ensures the decidability of other non-trivial properties of such systems, especially their alignment. Therefore, it would be computationally possible to verify its properties, although not necessarily tractable.

From a practical engineering perspective, while one may argue that real computers have limited memory and thus are always theoretically decidable, we argue that such constraint should be imposed over a tractable time such as $2^80$ rather than $2^1,000,000$ (1 megabit to enumerate the possible states). Similarly, the number of final states and loops should be such that it is possible for a group of human specialists to carefully analyze each of them, let us say, about $2^5$ final states and maximum loop length.

This requirement of always reaching a terminal state in a finite amount of (time)steps can also be expressed in terms of a Utility Function (outer-alignment) that penalizes the model for being in operation after the desired amount of time for halting. This penalization would have to be higher than the highest reward the model could ever receive. This would effectively reduce its capabilities in the time-domain.

Another implementation of this Utility Function would be a gradual change of the rewards and penalties with respect to the time(steps). This change would be inherently defined by the function itself.

\section*{Final Remarks}\label{sec:final}

The impossibility of having a general method that can assert whether an arbitrary AI is aligned or not does not mean that it is impossible to construct an AI that is provably aligned. Instead, it should be interpreted that there are many AIs that can not be proven to be aligned or not and that there is also a countable set of AIs that are proven to be aligned. Therefore, it is our objective to develop and utilize such a countable set of proven aligned AIs. The architecture and its development process are fundamental to ensure safety.

Developing an AI model that always halts allows for the alignment and other properties of the AI model to be asserted computationally, a task that would be computationally impossible for arbitrary models.

It is important to note that guaranteeing the decidability of the halting problem for a class of AI systems does not guarantee their alignment, it just guarantees that the alignment problem (and other properties) becomes decidable. This allows for further scrutiny of the model. This decidability is not a solution for the alignment problem, but it is a well-defined property and a desirable starting point for future works.

Although this article rephrases established theorems in the light of AI and does not provide an empirical contribution, it presents a compelling interpretation of the consequences of the alignment's undecidability and delineates a promising research direction for future works while also clarifying some misconceptions and pointing out other challenges to be addressed. We intend to apply the composition of aligned functions to generative models, namely, to LLMs.

\bibliography{references}

\section*{Author contributions statement}

G.A.M. proposed the main conceptual idea and wrote the initial draft.
M.R.O.A.M. contributed to the final interpretation and extended draft.
N.Y.S. and P.A.L.C. proposed a formal approach and discussed the main implications.
All authors reviewed the manuscript. 

\section*{Data availability statement}

The data that support the findings of this study are available on request from the corresponding author, G.A.M.

\section*{Competing interests}

The authors declare no competing interests.

\section*{Additional information}

\textbf{Correspondence} and requests for materials should be addressed to G.A.M.

\end{document}